\documentclass{article}
\usepackage{spconf,amsmath,graphicx}
\usepackage{enumitem}

\title{MADI: Inter-domain Matching and Intra-domain Discrimination for Cross-domain Speech Recognition}
\name{Jiaming Zhou\textsuperscript{1}\qquad Shiwan Zhao\sthanks{Independent researcher.}\qquad Ning Jiang\textsuperscript{2}\qquad  Guoqing Zhao\textsuperscript{2}\qquad  Yong Qin\textsuperscript{1}\sthanks{Corresponding author. This work was supported in part by NSF China (Grant No. 62271270), Tianjin Media Computing Center, Big Data Institute of Nankai University, and Mashang Consumer Finance.}}
\address{\textsuperscript{1} Nankai University, Tianjin, China \\
\textsuperscript{2} Mashang Consumer Finance Co., Ltd.}

\begin{document}
%\ninept
%
\maketitle
\begin{abstract}
End-to-end automatic speech recognition (ASR) usually suffers from performance degradation when applied to a new domain due to domain shift. Unsupervised domain adaptation (UDA) aims to improve the performance on the unlabeled target domain by transferring knowledge from the source to the target domain. To improve transferability, existing UDA approaches mainly focus on matching the distributions of the source and target domains globally and/or locally, while ignoring the model discriminability. In this paper, we propose a novel UDA approach for ASR via inter-domain \textbf{MA}tching and intra-domain \textbf{DI}scrimination (\textbf{MADI}), which improves the model transferability by fine-grained inter-domain matching and discriminability by intra-domain contrastive discrimination simultaneously. Evaluations on the Libri-Adapt dataset demonstrate the effectiveness of our approach. MADI reduces the relative word error rate (WER) on cross-device and cross-environment ASR by 17.7\% and 22.8\%, respectively.
\end{abstract}
\section{Introduction}
\label{sec:intro}
%With the rapid development of deep learning, end to end (E2E) Automatic Speech Recognition (ASR) models have made significant progress in a wide range of speech corpora\cite{ShigekiKarita2019ACS}. Existing approaches are based on the assumption that the training data and test data are from the same distribution. However, in reality the models trained on one dataset (source domain) do not perform well on another domain (target domain) when there are mismatches between training and test data caused by noises,devices,accents and so on. For instance, a ASR system trained on data recorded on Matrix Voice suffer from severe performance degradation on PlayStation Eye recordings. It's expensive and time-consuming  to collect sufficient data in target domains. Thus it is necessary to find an effective unsupervised  domain adaptation  method to improve the existing ASR models  on unlabeled target domain.  
Recent years have witnessed great progress in end-to-end automatic speech recognition (ASR) based on deep learning methods \cite{li2022recent}, which rely on large-scale labeled datasets and assume that training and testing data come from the same distribution. Nevertheless, when the models are trained on one domain (source domain) and tested on another domain (target domain), the performance degrades severely due to cross-domain distribution shift (domain shift). The causes of domain shift include variabilities of the acoustic environment, device, accent, and so on. Collecting sufficient labeled data for each target domain to train a good ASR model is expensive and time-consuming. %Therefore, in this paper, we aim to find an effective unsupervised domain adaptation (UDA) method to improve the ASR performance on the unlabeled target domain by leveraging the rich labeled data from the source domain.

Unsupervised domain adaptation (UDA) has been proposed to improve the ASR performance on the unlabeled target domain by leveraging the label-rich source domain. To transfer knowledge from the source to the target domain, previous work mainly focuses on matching the distributions of the source and target domains by learning domain-invariant representations. Generative adversarial nets (GAN) \cite{YiChenChen2019AIPNetGA,hu2018duplex} and domain adversarial learning technique \cite{SiningSun2018DomainAT,das2021best,HuHu2020REDATAR,DominikaWoszczyk2020DomainAN,huang22b_interspeech} have shown to be effective for global domain matching. To name a few, Chen et al. \cite{YiChenChen2019AIPNetGA} attempt to disentangle accent-specific and accent-invariant characteristics to build a unified end-to-end ASR system based on GAN. Sun et al. \cite{SiningSun2018DomainAT} propose domain adversarial training (DAT) to encourage the model to learn domain-invariant representations. % where they use an adversarial multi-task to predict which domain the frame is from and adversarially train the feature generation neural network.
Discrepancy-based methods, such as maximum mean discrepancy (MMD) \cite{gretton2012kernel} and correlation (CORAL) \cite{sun2016deep}, have recently been used to minimize feature distribution discrepancy between domains. 

More recently, local domain matching approaches have become popular for fine-grained distribution matching, which aligns the distributions of the relevant subdomains across different domains. Hu et al. \cite{HangRuiHu2022ClassAwareDA} propose subdomain distribution matching to extract domain-invariant embeddings for speaker verification. In CMatch \cite{WenxinHou2021CrossdomainSR}, a character-level distribution matching method is adopted to address domain shift. 
The inter-domain matching, either globally or locally, improves the model transferability. However, simply pushing the source and target domains together may compromise the discriminability of the model in the target domain \cite{chen2019transferability, Yang2020MindTD}.

\begin{figure*}[t]
\centering
\includegraphics[width=0.96\textwidth]{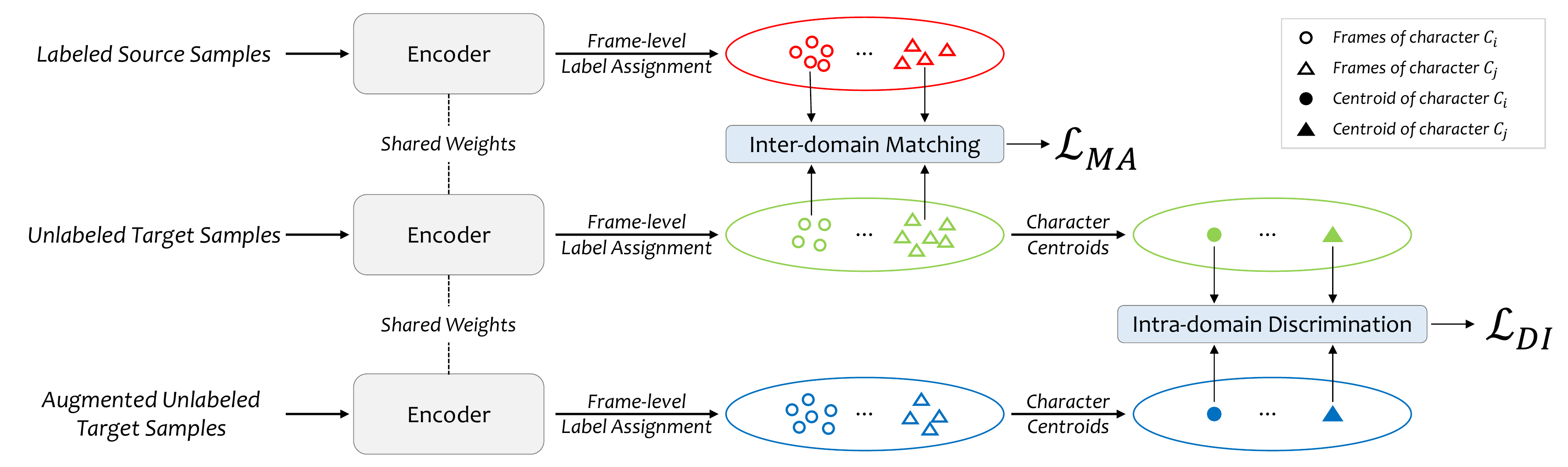}
\caption{Overview of our MADI framework. For source samples with ground-truth labels and (augmented) target samples with pseudo labels, we use encoders to extract features, and then employ the CTC decoder to assign labels to the encoded frames. Then we compute the two adaptation losses: $\mathcal{L}_{MA}$ for inter-domain matching and $\mathcal{L}_{DI}$ for intra-domain discrimination. The former matches the character-level distributions between the source and target domains. The latter employs the contrastive learning method to push the centroids of different characters away from each other in the target domain. For concise, the joint CTC-Attention loss $\mathcal{L}_{ASR}$ is omitted in the figure.}
\label{network}
\end{figure*}

To address the aforementioned issue, in this paper, we propose MADI, a novel UDA approach for ASR via inter-domain matching and intra-domain discrimination. With fine-grained inter-domain matching, the proposed method improves the model transferability, while with intra-domain contrastive discrimination, we enhance the model discriminability in the target domain. %compared to conventional UDA methods. %Intuitively, cross-domain contrastive learning approaches \cite{Huang_2022_CVPR,NEURIPS2021_288cd256} from computer vision domain can be leveraged for this purpose, which constructs positive and negative pairs from both domains.
Specifically, our framework contains two main components for domain adaptation (see figure \ref{network}). Firstly, inspired by CMatch \cite{WenxinHou2021CrossdomainSR}, we employ an inter-domain matching component that matches the character-level distributions between the labeled source domain and unlabeled target domain. Secondly, motivated by the success of contrastive learning \cite{chen2020simple, jiang2020speech}, we generate augmented unlabeled target data and then propose an intra-domain discrimination component to ensure that the centroids of the same characters are pulled closer, while the centroids of different characters are pushed apart in the target domain. This facilitates learning discriminative representations for the unlabeled target domain. The above two components are jointly optimized using the well-defined MMD \cite{gretton2012kernel} and normalized temperature-scaled cross-entropy loss (NT-Xent) \cite{chen2020simple}. 

We note that cross-domain contrastive learning approaches \cite{Huang_2022_CVPR,NEURIPS2021_288cd256} from the computer vision domain are similar to our work, which performs inter-domain alignment and intra-domain discrimination by deliberately constructing positive and negative pairs. Directly adopting such contrastive learning approaches at the frame level results in an explosion of computational complexity, while at the character-prototype level it leads to sub-optimal performance (see Section \ref{sec:result}). 

We conduct extensive experiments on the Libri-Adapt dataset \cite{AkhilMathur2020LibriAdaptAN}. The results demonstrate the effectiveness of our approach. MADI outperforms the state-of-the-art UDA methods and achieves a relative performance improvement of 17.7\% and 22.8\% word error rate (WER) on cross-device and cross-environment ASR, respectively.

\section{Our Method}
\label{sec:method}

%Unsupervised cross-domain ASR adaptation is aiming to  leverage the available labeled source-domain data to improve the robustness of ASR on the unlabeled target-domain.  Our key idea is to match the inter-domain character-wise distribution to enhance the model transferability, and discriminate intra-domain features of the same character to improve discrimination capability on target domain. The structure of the proposed  Inter-domain \textbf{MA}tching and Intra-domain \textbf{DI}scrimination adaptation method is shown in figure \ref{network}. The overall loss has three parts that we will introduce separately.
UDA for ASR aims to exploit labeled source data $(X_S, Y_S)$ to improve the ASR performance on unlabeled target data $X_T$. Our key idea is to match the character-level distributions between the source and target domains to enhance the model transferability and push the features of different characters apart to improve the model discriminability in the target domain. The overall structure of the proposed adaptation method, inter-domain matching and intra-domain discrimination (MADI), is shown in figure \ref{network}. Before describing the two adaptation components in detail, we briefly introduce the basic ASR model used in our framework.

\subsection{Basic ASR Model}
We build a joint CTC-Attention model following open-source Wenet \cite{ZhuoyuanYao2021WeNetPO}, which consists of three parts: shared encoder, CTC decoder, and attention decoder. The loss of ASR is as follows:
\begin{equation} 
\mathcal{L}_{ASR}(X,Y)={\lambda}\mathcal{L}_{CTC}(X,Y)+{(1-\lambda)}\mathcal{L}_{ATT}(X,Y),
\label{asr_loss}
\end{equation}
where $X$ and $Y$ are the acoustic input and corresponding labels, respectively. $\mathcal{L}_{CTC}$ is the CTC loss, and $\mathcal{L}_{ATT}$ is the attention loss. The hyperparameter ${\lambda}$ balances the two losses.

\subsection{Inter-domain Matching}
%Experiments show that the model transferability is improved when the extracted features of the same class from different domains are matched in the latent space, therefore inter-domain matching is proposed. 
%Inspired by CMatch\cite{WenxinHou2021CrossdomainSR}, to achieve frame-level label assignment by  pre-trained model with  $(X_S,Y_S)$, we employ CTC pseudo label of $n$-th frame by,
Since our inter-domain matching is based on character level, we first assign labels to frames, which is time-consuming. %Given labeled source data $(X_S,Y_S)$ and unlabeled target data $X_T$, 
We follow CMatch \cite{WenxinHou2021CrossdomainSR} to achieve efficient and accurate frame-level label assignment. With the pre-trained model on $(X_S,Y_S)$, we obtain the CTC \cite{graves2006connectionist} pseudo label of the $n$-th frame by
\begin{equation}
\hat{Y}_n=arg \mathop{max}\limits_{Y_n}P_{CTC}(Y_n|X_n).
\label{ctc_loss}
\end{equation}

After acquiring labels for each encoded frame, the conditional distributions $P(Y|X)$ of each character in both the source and target domains could be obtained. 
%MMD\cite{gretton2012kernel} has been wisely used to address distribution alignment in transfer learning tasks, therefore we employ it to match the conditional distribution between the same character. 
We then adopt the widely used MMD distance \cite{gretton2012kernel} to match the conditional distributions between the same characters across domains. The inter-domain matching loss $\mathcal{L}_{MA}$ is as follows:
\begin{equation}
\mathcal{L}_{MA}=\frac{1}{N}\sum_{i}^NMMD(\mathcal{H}_k,X_S^{C_i},X_T^{C_i}),
\label{MA_loss}
\end{equation}
where $N$ is the total number of characters. $C_i$ means the $i$-th character of the symbol set $C$. $\mathcal{H}_k$ is the reproducing kernel Hilbert space, and $k$ is the Gaussian kernel function we adopted.
The model is trained to minimize $\mathcal{L}_{MA}$ to match the distributions of the same characters across domains.

\subsection{Intra-domain Discrimination}
%Inter-domain matching could improve the model transferability, but accompanied with the model discrimination capability deterioration. Therefore, we introduce intra-domain discrimination to overcome this shortcoming. 
We employ the contrastive learning approach \cite{chen2020simple} in the target domain to improve the model discriminability. Given target domain data $X_T$, we first generate an augmented version $X_{aug}$ by pitch randomization, reverberation, and temporal masking. And then, similar to inter-domain matching, we achieve frame-level label assignment using CTC pseudo labels for both $X_T$ and $X_{aug}$. Adopting contrastive learning at the frame level will result in an explosion of computational complexity, so we apply the contrastive learning on the character prototypes by computing the centroids for $2*N$ symbols in a batch of $X_T$ and $X_{aug}$. 
The character centroids of the same symbol form positive pairs $(\widetilde{X}_T^{C_i},\widetilde{X}_{aug}^{C_i})$, and the rest ones from $X_T$ and $X_{aug}$ are counted as negative pairs.
We attempt to keep positive pairs together and push negative pairs apart in the batch by minimizing the modified NT-Xent loss \cite{chen2020simple}. The intra-domain discrimination loss $\mathcal{L}_{DI}$ is defined as: 
% \begin{equation}
%     \mathcal{L}_{DI}=\sum_i^B\sum_j^B
% \end{equation}
% \begin{equation}
% \mathcal{L}_{DI}(X^{C_i},X_{aug}^{C_i})\\
% =-log\frac{sim(F(X^{C_i}),F(X_{aug}^{C_i}))/\tau}
% {sim(F(X^{C_i}),F(X_{aug}^{C_i}))/\tau+\sum_{j\neq i}{sim(F(X^{C_i}),F(X_{aug}^{C_j}}))/\tau}
% \label{DI_loss}
% \end{equation}
\begin{equation}\label{DI_loss}
\begin{split}
&\mathcal{L}_{DI}(\widetilde{X}^{C_i}_T,\widetilde{X}_{aug}^{C_i})\\
&=-log\frac{\psi(\widetilde{X}^{C_i}_T,\widetilde{X}_{aug}^{C_i})}
        {\psi(\widetilde{X}^{C_i}_T,\widetilde{X}_{aug}^{C_i})+\sum_{j\neq i \atop d\in\{T,aug\}}
        {\psi(\widetilde{X}^{C_i}_T,\widetilde{X}_{d}^{C_j})}},
\end{split}
\end{equation}
where $1 \leq i,j \leq N$, and $\psi(a,b)=exp(sim(f(a),f(b))/\tau)$.
$sim(u,v)$ = $\frac{u^Tv}{\Vert u \Vert _2 \Vert v \Vert _2}$ denotes the cosine similarity of $u$ and $v$. $f()$ means features extracted by the encoder, and $\tau$ is the temperature hyperparameter. Note that $\mathcal{L}_{DI}$ is the average of $\mathcal{L}_{DI}(\widetilde{X}^{C_i}_T,\widetilde{X}_{aug}^{C_i})$ and $\mathcal{L}_{DI}(\widetilde{X}_{aug}^{C_i}, \widetilde{X}^{C_i}_T)$ for all positive pairs.

\subsection{Overall Loss}
The overall loss function includes the joint CTC-Attention loss, the inter-domain matching loss, and the intra-domain discrimination loss, which is defined as follows:
\begin{equation} 
\mathcal{L}=\mathcal{L}_{ASR}+\alpha*\mathcal{L}_{MA}+\beta*\mathcal{L}_{DI},
\label{all_loss}
\end{equation}
where $\alpha$ and $\beta$ are hyperparameters to tradeoff the impact of $\mathcal{L}_{MA}$ and $\mathcal{L}_{DI}$.

\section{Experimental setup}
\label{sec:exp}
\subsection{Dataset}
Our experiments are conducted on the Libri-Adapt dataset \cite{AkhilMathur2020LibriAdaptAN} which is derived from Librispeech-clean-100. The Libri-Adapt provides 72 different domains for domain adaptation study, which is recorded under 4 background noise conditions from 3 speaker accents on 6 different embedded microphones. In this work, we focus on cross-device and cross-environment adaptation, i.e., source-domain and target-domain data are recorded by different devices or in different environments. 

Recent research \cite{AkhilMathur2019Mic2MicUC} has shown that the variabilities of microphones across different devices significantly influence their outputs. 
In this paper, we employ 3 parts of the Libri-adapt dataset for cross-device experiments including Matrix Voice (M), Respeaker (R), and PlayStation Eye (P).
Matrix Voice and Respeaker are circular 7-channel microphone arrays integrated with acoustic signal processing algorithms while Playstation-Eye is a 4-channel microphone for voice interactive games. 
For cross-environment ASR adaptation, we select clean Respeaker as the source domain and 3 types of background noise including Rain, Wind, and Laughter as different target domains.
%Each two of them can form a cross-device adaptation task, leading to 6 tasks totally. % When we select one of them as target domain for unsupervised domain adaptation
%Note that we do not use any labels from the target domain except for evaluation. 
In our experiments, we do not use any labels from the target domain during training and randomly split 10\% utterances in the source domain as the validation set.

\subsection{Baselines}
The following methods are considered for comparison:

\begin{itemize}[leftmargin=*]

\item {\bf SO}: The source-only (SO) method trains the ASR model on the source domain and directly applies it to the target domain without adaptation.

\item {\bf DAT} \cite{SiningSun2018DomainAT}: Domain adversarial training (DAT) is a popular UDA method which adversarially trains a discriminator and an encoder to encourage the encoder to learn domain-invariant features. We re-implemented DAT with a domain discriminator consisting of fully-connected linear layers. 
%A domain classifier is introduced to discriminate which domain features extracted by encoder are from. Further, apply gradient reverse layer before the domain classifier to encourage the encoder to learn domain-invariant features. We re-implemented DAT with a domain discriminator consisting of fully-connected linear layers.

\item {\bf CMatch} \cite{WenxinHou2021CrossdomainSR}: It is a character-level distribution matching method, which employs CTC pseudo labels to achieve frame-level label assignment and then reduces the character-level distribution divergence between the source and target domains using MMD. We also re-implement CMatch.

\item {\bf CDCL}: Cross-domain contrastive learning (CDCL) approaches are popular in the computer vision domain \cite{Huang_2022_CVPR,NEURIPS2021_288cd256}. We implement the idea at the character prototype level by considering the centroids of the same characters from different domains as positive pairs and the centroids of different characters from both domains as negative pairs.

\end{itemize}

\subsection{Implementation Details}
For fair comparison, we implement all baselines and our method based on Wenet \cite{ZhuoyuanYao2021WeNetPO} codebase.  All experiments use 80-dimensional log Mel-filter banks (FBANK) features with a 25ms window and a 10ms shift. 
The underlying transformer model has 12 encoder layers and 6 decoder layers. Both of them have 4 attention heads and 2048 linear units. 
The CTC loss weight $\lambda$ is set to 0.3 following \cite{ZhuoyuanYao2021WeNetPO}. 
The hyperparameters $\alpha$ and $\beta$ are set to 5 in our method. %and the UDA loss weight of the other methods is set to 10. 
The temperature $\tau$ is 0.1 during our experiment. 
The training data of the target domain is augmented using the open-source tool WavAugment \cite{wavaugment2020} by pitch randomization, reverberation, and temporal masking. 
When training, we filter out utterances over 17.5s. We employ the learning rate from  $1\times10^{-3}$ to $8\times10^{-3}$ and adam optimizer with a learning rate schedule including 25,000 warm-up steps. 
Beam size is 10 for decoding. 
The batch size is 64 and the epoch is 150/180 for cross-device/-environment models. 
The output dimension is 31 consisting of 26 letters and 5 symbols.  
The attention-rescoring mode we adopt at the testing time always keeps the best performance among the 4 decoding methods provided by the model.

\section{Results}
\label{sec:result}
% the word error rate (WER) of in-domain results is in Table\ref{baseline} which the model is trained and tested in source domain.
We first report the WER results of the standard ASR in Table \ref{indomain} for comparison. The standard ASR is an in-domain model with training and testing data from the same domain.

\begin{table}[t]
\caption{WER on standard ASR}
\center
\label{indomain}
\begin{tabular}{cc}
\hline
Domain               & WER \\ \hline
Matrix Voice (M)     & 23.74        \\
Play Station Eye (P) & 20.77        \\
ReSpeaker (R)        & 22.82        \\ \hline
Average             & 22.44        \\ \hline
\end{tabular}
\end{table}

\subsection{Cross-device Adaptation}
The main results of cross-device ASR are reported in Table \ref{cross-device}. The task name indicates the source and target domains, e.g., M$\to$P denotes the source domain Matrix Voice (M) and the target domain Playstation Eye (P).

%It can be seen that the performance of SO model deteriorates greatly due to domain mismatching. DAT and CMatch which use global domain and local domain alignment strategies respectively get improvements on all tasks compared with SO. CDCL is superior to DAT but inferior to CMatch, indicating that CDCL could not achieve character level alignment as effective as CMatch due to pushing positive pairs together and negative pairs apart simultaneously. MADI achieves the lowest WER on 5 tasks and the best mean performance among all methods exceeding SO by 17.7\% relatively. The results in Table \ref{cross-device} indicate that matching domain distribution globally and locally simply may compromise the model discriminability and it is effective to employ inter-domain matching and inter-domain discrimination simultaneously for cross-device ASR adaptation.
Firstly, we observe that the performance of SO is severely degraded due to domain mismatch. Both DAT and CMatch improve performance on all tasks through inter-domain matching. CMatch outperforms DAT, indicating that fine-grained local domain alignment is superior to global alignment. Secondly, CDCL is inferior to CMatch, although CDCL attempts to align positive pairs across domains and separate negative pairs at the same time. The reason is that the inter-domain matching ability of CDCL at the character prototype level is weaker than that of MMD used in CMatch.  Thirdly, MADI achieves the lowest WER on 5 of the 6 tasks and the best average performance. MADI significantly outperforms SO by 17.7\% relatively, which demonstrates the effectiveness of our approach by improving the transferability and discriminability simultaneously.

\begin{figure}[t]
%\label{scatter}
\begin{minipage}[b]{.48\linewidth}
  \centering
  \centerline{\includegraphics[width=4.5cm]{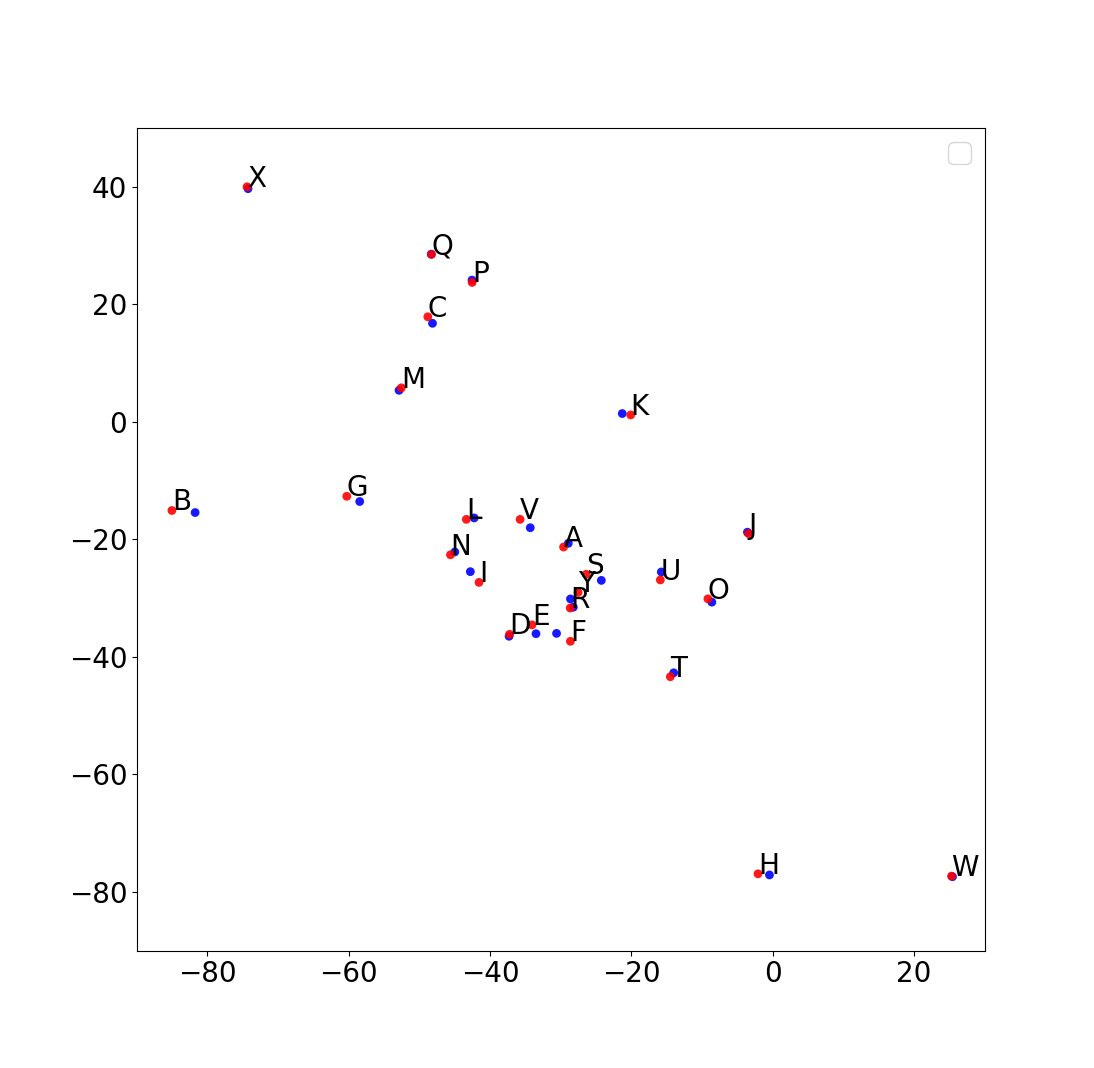}}
%  \vspace{1.5cm}
  \centerline{(a) CMatch}\medskip
\end{minipage}
\hfill
\begin{minipage}[b]{0.48\linewidth}
  \centering
  \centerline{\includegraphics[width=4.5cm]{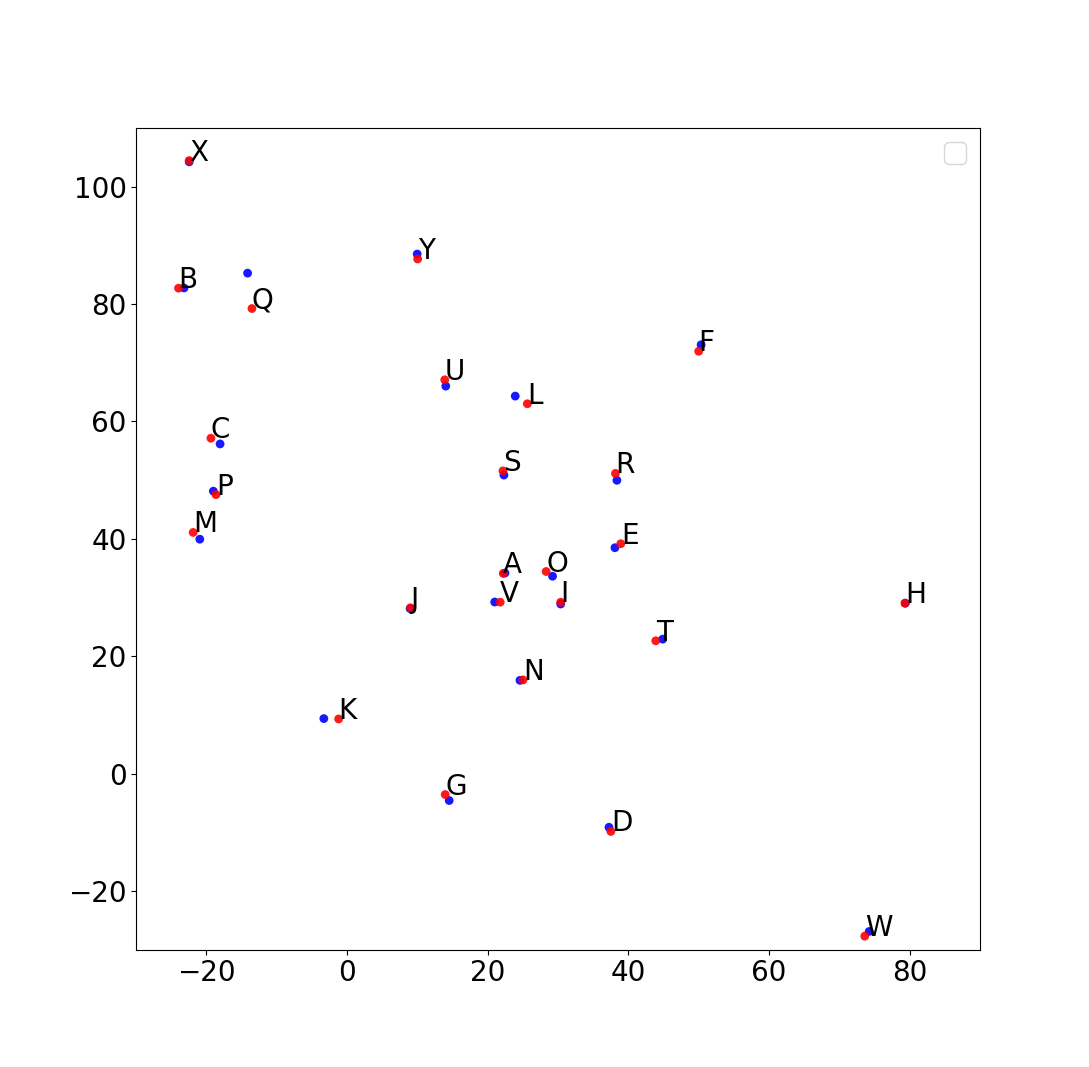}}
%  \vspace{1.5cm}
  \centerline{(b) MADI}\medskip
\end{minipage}
\caption{Compared to CMatch, feature centers of characters are more spread out in MADI. Dots in red and blue indicate the source and target domains, respectively.}
\label{fig:tsne}
\end{figure}

To further demonstrate the ability of MADI to enhance the model discriminability, we visualize the feature distributions of CMath and MADI in figure \ref{fig:tsne}. We observe that the centroids of the same characters from different domains are well aligned in CMatch while the distances between different characters are somewhat close. By applying contrastive discrimination in the target domain, characters in MADI are pushed away from each other, indicating the improvement of the model discriminability.

\subsection{Cross-environment Adaptation}
The cross-environment ASR results are shown in Tabel \ref{cross-environment}. 
%SO trained with Respeaker in the clean environment also suffers serious performance degradation due to domain mismatch. 
We also observe that MADI outperforms DAT and CMatch. Moreover, compared to SO trained with Respeaker in the clean environment, MADI reduces relative WER by 22.8\%, indicating its effectiveness for cross-environment adaptation. Note that our re-implementation of CMatch performs better than what the paper \cite{WenxinHou2021CrossdomainSR} reports.

% \subsection{Ablation Study}
% To contrast the effect of each component of MADI, we conduct several experiments including DI-only, MA-only and DI-DI. Note that DI-only employs contrastive learning discrimination between source and target domains over baseline, while MA-only uses inter-domain matching merely. Moreover, DI-DI increases intra-domain discrimination based on DI-only. 

% \begin{figure*}[t]
% \includegraphics[width=0.5\textwidth]{test.pdf}
% \caption{}
% \label{network}
% \end{figure*}

% \begin{figure*}
% \includegraphics[width=0.5\textwidth]{scatter.jpg}
% \caption{}
% \label{network}
% \end{figure*}

% \begin{table}[]
% \begin{tabular}{ccccccc}
% \label{all_res}
% \hline
% Task       & Source-only & DAT   & DI-only & DI-all & Cmatch         & Our method     \\ \hline
% M$\to$P    & 23.94       & 21.92 & 21.53   & 22.03  & 20.28          & \textbf{20.25} \\
% M$\to$R    & 26.43       & 23.58 & 24.49   & \      & 22.79          & \textbf{22.61} \\
% P$\to$M    & 28.97       & 25.02 & 24.99   & \      & 23.91          & \textbf{23.41} \\
% P$\to$R    & 23.54       & 22.64 & 22.06   & \      & 20.25          & \textbf{19.7}  \\
% R$\to$M    & 34.95       & 28.34 & 29.45   & \      & 28.68          & \textbf{27.27} \\
% R$\to$P    & 22.7        & 21.84 & 20.53   & \      & \textbf{18.82} & 18.89          \\\hline
% Average    & 26.76       & 23.89 & 23.84   & \      & 22.46          & \textbf{22.02}\\ \hline
% \end{tabular}
% \end{table}

\begin{table}[t!]
\center
\caption{WER on cross-device ASR}
\label{cross-device}
\begin{tabular}{ccccccc}\hline
Task             & SO         & DAT          & CMatch        & CDCL    & MADI     \\ \hline
M$\to$P          & 23.94            & 21.92        & 20.28          &21.53  & \textbf{20.25}      \\
M$\to$R          & 26.43            & 23.58        & 22.79          &24.49  & \textbf{22.61}      \\
P$\to$M          & 28.97            & 25.02        & 23.91          &24.99  & \textbf{23.41}      \\
P$\to$R          & 23.54            & 22.64        & 20.25          &22.06  & \textbf{19.7}       \\
R$\to$M          & 34.95            & 28.34        & 28.68          &29.45  & \textbf{27.27}      \\
R$\to$P          & 22.7             & 21.84        & \textbf{18.82} &20.53  & 18.89               \\ \hline
Average       & 26.76               & 23.89        & 22.46          &23.84  & \textbf{22.02}      \\ \hline
\end{tabular}
\end{table}

\begin{table}[t!]
\center
\caption{WER on cross-environment ASR}
\label{cross-environment}
\begin{tabular}{cclcc}\hline
Task            & SO    & DAT     & CMatch              & MADI \\\hline
Rain            & 33.06       & 33.64   & 26.24             & \textbf{25.82} \\
Wind            & 26.19       & 27.17   & 21.40            & \textbf{21.06} \\
Laughter        & 31.12       & 28.52   & 23.48             & \textbf{22.91} \\\hline
Average         & 30.12       & 29.78   & 23.71             & \textbf{23.26} \\\hline
\end{tabular}
\end{table}

% \begin{table}[h!]
% \caption{Comparison of different MADI combinations}
% \begin{tabular}{ccccc}
% \hline
% Task       & CL         & MA-only       & CL-DI     & MADI(ours)     \\\hline
% M$\to$P    & 21.53      & 20.28         & 22.03     & \textbf{20.25}  \\
% M$\to$R    & 24.49      & 22.79         & 24.03     & \textbf{22.61} \\
% P$\to$M    & 24.99      & 23.91         & 25.40     & \textbf{23.41}  \\
% P$\to$R    & 22.06      & 20.25         & 22.52     & \textbf{19.7}  \\
% R$\to$M    & 29.45      & 28.68         & 28.26     & \textbf{27.27} \\
% R$\to$P    & 20.53      & \textbf{18.82}& 21.28     & 18.89          \\\hline
% Average    & 23.84      & 22.46         & 23.92     & \textbf{22.02} \\\hline
% \end{tabular}
% \end{table}

\section{Conclusion}
\label{sec:conclusion}
In this paper, we propose an unsupervised cross-domain ASR adaptation method via inter-domain matching and intra-domain discrimination. Our approach improves the model transferability and discriminability simultaneously. Experimental results on the Libri-Adapt dataset demonstrate the effectiveness of our approach.

\vfill\pagebreak
\bibliographystyle{IEEEbib}
\bibliography{strings,refs}
\end{document}